\documentclass[sigconf]{acmart}
\setcopyright{acmcopyright}
\usepackage[export]{adjustbox}[2011/08/13]
\begin{document}
\fancyhead{}
\copyrightyear{2019} 
\acmYear{2019} 
\acmConference[AVEC '19]{9th International Audio/Visual Emotion Challenge and
Workshop}{October 21, 2019}{Nice, France}
\acmBooktitle{9th International Audio/Visual Emotion Challenge and Workshop (AVEC '19),
October 21, 2019, Nice, France}
\acmPrice{15.00}
\acmDOI{10.1145/3347320.3357697}
\acmISBN{978-1-4503-6913-8/19/10}

\title{Multi-level Attention network using text, audio and video for Depression Prediction}

\author{Anupama Ray}
\affiliation{
  \institution{IBM Research, India}
  }
  \orcid{0002-9193-5017}
\email{anupamar@in.ibm.com}

\author{Siddharth Kumar}
\affiliation{
  \institution{IIIT Sricity, India}
  }
\email{siddharth.k16@iiits.in}

\author{Rutvik Reddy}
\affiliation{
 \institution{IIIT Sricity, India}
 }
\email{rutvikreddy.v16@iiits.in}


\author{Prerana Mukherjee}
\affiliation{
  \institution{IIIT Sricity, India}
  }
\email{Prerana.m@iiits.in}

\author{Ritu Garg}
\affiliation{
  \institution{Intel Corporation}
   }
\email{ritu@intel.com}

\renewcommand{\shortauthors}{Anupama Ray et al.}

\begin{abstract}
Depression has been the leading cause of mental-health illness worldwide. Major depressive disorder (MDD), is a common mental health disorder that affects both psychologically as well as physically which could lead to loss of lives. Due to the lack of diagnostic tests and subjectivity involved in detecting depression, there is a growing interest in using behavioural cues to automate depression diagnosis and stage prediction. The absence of labelled behavioural datasets for such problems and the huge amount of variations possible in behaviour makes the problem more challenging. This paper presents a novel multi-level attention based network for multi-modal depression prediction that fuses features from audio, video and text modalities while learning the intra and inter modality relevance. The multi-level attention reinforces overall learning by selecting the most influential features within each modality for the decision making. We perform exhaustive experimentation to create different regression models for audio, video and text modalities. Several fusions models with different configurations are constructed to understand the impact of each feature and modality. We outperform the current baseline by 17.52\% in terms of root mean squared error. 
\end{abstract}



\maketitle

\section{Introduction} 
Depression is a one of the common mental health disorders and according to WHO, 300 million people around the world have depression \footnote{{\tiny \url{https://www.verywellmind.com/depression-statistics-everyone-should-know-4159056}}}. It is a leading cause of mental disability, has tremendous psychological and pharmacological affects and can in the worst case lead to suicides.
A big barrier to effective treatment of MDD and its care is inaccurate assessment due to the subjectivity involved in the assessment procedure. Most assessment procedures rely on using questionnaires such as Physical Health Questionnaire Depression Scale (PHQ), the Hamilton Depression Rating Scale (HDRS), or the Beck Depression Inventory (BDI) etc. All of these questionnaries used in screening involve patient's response which is often not very reliable due to different subjective issue of an individual. The symptoms of MDD are covert and there could be some individuals who complain a lot in general even without having mild depression, whereas most severely depressed patients do not speak much in the screening test. Thus, it is very challenging to diagnose early depression and often people are misdiagnosed and prescribed antidepressants. Unlike physical ailments, there are no straightforward diagnostic tests for depression and clinicians have to routinely screen individuals to determine whether the type of clinical or chronic depression. Studies have shown that around 70\% sufferers from MDD have consulted a medical practitioner \cite{stats}. Most practitioners follow the gold standard Physical Health Questionnaire \cite{PHQ-8}, which has questions to check for symptoms such as fatigue, sleep struggles, appetite issues etc. Diagnosis is based on the judgement of the practitioner (which could be biased from past-education or past experience). Often there are false positives or false negatives with the PHQ screening which lead to misjudgement in diagnosis. 

The huge need for depression detection and the challenges involved motivated the affective computing research community to use behavioural cues to learn to predict depression, Post-traumatic stress disorder, and related mental disorders \cite{AVEC2016}. Behavioral cues such as facial expression, prosodic features from speech have proven to be excellent features for depression prediction \cite{8,58}. 

In this paper, we present a novel framework that invokes attention mechanism at several layers to identify and extract important features from different modalities to predict level of depression. The network uses several low-level and mid-level features from both audio and video modalities and also sentence embeddings on the speech-to-text output of the participants. We show that attention at different levels gives us the ratio of importance of each feature and modality, leading to better results. We perform several experiments on each feature from different modality and combine several modalities. Our best performing network is the all-feature fusion network which outperforms the baseline by 17.52\%.
The individual feature-based attention network outperforms the baseline by 20.5\% and the attention based text model outperforms the state-of-art by 8.95\%, the state-of-art network being an attention based text transcription network as well \cite{qureshi2019verbal}. 

\subsection{Contributions}
The key contributions of this work is given as follows:
\begin{itemize}
    \item Attention based fusion network: We present a novel feature fusion framework that utilizes several layers of attention to understand importance of intra-modality and inter-modality features for the prediction of depression levels.
    \item The proposed approach outperforms the baseline fusion network by - on root mean square error.
    \item An improved attention based network trained on all three modalities which outperforms the baseline by 17.52\%.
    
\end{itemize}

The remaining paper is organized as follows: Section \ref{sec:related} presents the state-of-art methods for depression classification. We present a brief overview about the proposed multi-level attention network in section \ref{sec:dataset}, followed by a brief of the dataset used. In section   \ref{sec:methodology}, the detailed methodology for each model built on individual features or fusion is described. Section \ref{sec:results} explains the results of all the models and presents all ablation studies followed by discussions and future work in section \ref{sec:conclusion}.

\section{Related Work}
\label{sec:related}
In this section, we briefly provide a review of the various works done in context of distress analysis using multimodal inputs such as text, speech, facial emotions and multimodal sentiment analysis.

\subsection{Depression detection from Speech}
Speech, more specifically non-verbal paralinguistic cues have gained significant popularity in distress prediction and similar tasks due to two main reasons. First, clinicians use speech traits such as diminished prosody, less or monotonous verbal activity production and energy in speech as important markers in the diagnosis of distress. Secondly, speech being an easy signal to record (non-invasive and non-intrusive), makes it the best candidate for all automation tasks \cite{cummins2015review}. 
Cummins et.al. \cite{cummins2015review} provide an exhaustive review of depression and suicide risk assessment using speech analysis. They investigate the usage of vocal bio- markers to associate clinical scores in case of depression signs.
In \cite{stasak2016investigation}, authors perform distress assessment in speech signals to infer emotional information expressed while speaking. This amounts to quantifying various expressions such as anger, valence, arousal, dominance etc. In \cite{mitra2016noise}, authors provide a comparative study on noise and reverberation on depression prediction using mel-frequency cepstral coefficients (MFCCs) and  damped oscillator cepstral coefficients (DOCCs). Cummins et.al. in \cite{cummins2015analysis} investigate change in spectral and energy densities of speech signals for depression prediction. They analyze the acoustic variability in terms of weighted variance, trajectory of speech signals and volume to measure depression. The cross-cultural and cross-linguistic characteristics in depressed speech using vocal biomarkers is studied in \cite{alghowinem2016cross}. 
In \cite{williamson2016detecting}, authors study the neurocognitive changes influencing the dialogue delivery and semantics. Semantic features are encoded using sparse lexical embedding space and context is drawn from subject's past clinical history. 

\subsection{Facial Emotion Analysis}
Although the inherent relationship between verbal content and mental illness level is more prominent, the visual features also play a pivotal role to reinstate the deep association of depression to facial emotions. It has been observed that patients suffering from depression often have distorted facial expressions for e.g. eyebrow twitching, dull smile, frowning faces, aggressive looks, restricted lip movements, reduced eye blinks etc. With the quantum of proliferating video data and availability of high end built-in cameras in wearables and surveillance sectors, analyzing the facial emotions and sentiments is the growing trend amongst the vision community. In \cite{poria2016convolutional}, authors utilize convolutional multiple kernel learning approaches for emotion recognition and sentiment analysis in videos. Dalili et.al. conducted a thorough study for meta-analysis of the association between existing facial emotion recognition and depression \cite{dalili2015meta}. In \cite{poria2017context}, authors rely on temporal LSTM based techniques for capturing the contextual information from videos in sentiment analysis task. Valstar et.al. introduced Facial Expression Recognition and Analysis challenge (FERA 2017) dataset \cite{valstar2017fera} to estimate the head pose movements and identify the action units against them. It requires to quantify the facial expressions in such challenging scenarios. Ebrahimi et.al. introduced Emotion Recognition in the Wild (EmotiW) Challenge dataset \cite{ebrahimi2015recurrent} and utilize a hybrid convolutional neural network-recurrent neural network (CNN-RNN) framework  for facial expression analysis. These datasets have been very crucial in advancing state-of-art in research around facial expression recognition and distress prediction. In \cite{baltruvsaitis2016openface}, authors present OpenFace an open source interactive tool to estimate the facial behaviour. This is a widely popular tool and in this paper, we have used the features extracted from OpenFace only as the video low-level features. OpenFace gives us features for face landmark regions, head pose estimation and eye gaze estimation and converts it into a reliable facial action unit. In \cite{mccleery2015meta}, authors investigate the meta-analysis of facial gestures to identify the schizophrenia based event triggers. In \cite{graziano2016attention}, authors provide a meta-analysis on attention deficit hyperactivity disorder (ADHD) and dysregulation of children's emotions. They provide an attempt to establish a coherent link between ADHD and emotional reactions. 

\subsection{Depression Quotient Detection in Text}
Along with video and audio, the verbal content of what a person speaks is critically important to be able to diagnose depression and stress. With the surge of social media usage, there is a lot of textual data inflow from social media which has given researchers the opportunity to try to analyze distress from text.
Such data could help in sentiment analysis and provide insights to sudden aberrations in the personality traits of the user as reflected in one's posts. In \cite{shen2017depression}, authors leverage social media platforms to detect depression by harnessing the social media data. They categorize the tweets gathered from Twitter API into depression and non-depression data. They extract various feature groups correlated to six depression levels and further utilize the multimodal depressive dictionary learning for online behaviour prediction of the Twitter user base. In \cite{cavazos2016content}, authors have inspected the tweet content ranging from common themes to trivial mention of depression to classify them into relevant category of distress disorder levels. In \cite{hassan2017sentiment}, authors present a sentiment analysis framework using social media data and mine patterns based on emotion theory concepts and natural language processing techniques. Ansari et.al. propose a Markovian model to detect depression utilizing content rating provided by human subjects \cite{ansari2018dcr}. The users are presented with series of contents and then asked to rate them based on the reactions tapped and tendency to skip it the depression level is associated to the event. In \cite{jia2018mental}, authors examine the onset of triggered events for mental illness specifically stress and depression based on social media data encompassing different ethnic backgrounds. Tong et.al. \cite{tong2019inverse} utilize a novel classifier inverse boosting pruning trees to mine online social behaviour. This enables depression detection at early stages. In \cite{hao2019providing}, authors adopt clustering techniques to quantify the anxiety and depression indices on questionnaire textual data. Further, the correlation amongst anxiety, depression and social data is investigated.For most of social media data, the text analysis is done for short text and these classifiers dont work well in a conversation setting which happens during the counselling/screening sessions.


\subsection{Multimodal Approaches for Distress Detection}
In \cite{morales2018linguistically}, authors provide a comprehensive review on the fusion techniques for depression detection. They also propose a computational linguistics based fusion approach for multimodal depression detection. In \cite{lam2019context}, authors analyse the depression levels from clinical interview corpus DAIC dataset based on context aware feature generation techniques and end-to-end trainable deep neural networks. They further infuse data augmentation techniques based on topic modelling in the transformer networks. Zhang et.al. released a multimodal spontaneous emotion corpus for human behaviour analysis \cite{zhang2016multimodal}. Facial emotions are captured by 3D dynamic ranging, high resolution video capture and infrared imaging sensors. Apart from facial context, blood pressure, respiration and pulse rates are monitored to gauge the emotional state of a person. 
Using the data released in AVEC challenge \cite{AVEC2015}, audio, video and physiological parameters are investigated to observe key findings on emotional state of the subjects. In \cite{poria2016fusing}, authors fuse audio, visual and textual cues for harvesting sentiments in multimedia content. They utilize feature and decision level fusion techniques to perform affective computing. In \cite{alghowinem2016multimodal}, authors utilize paralinguistic, head pose and eye gaze fixations for multimodal depression detection. With the help of statistical tests on the selected features the inference engine will classify the subjects into depressed and healthy categories.  

When combining multiple modalities it is important to understand the contribution of each modality in the task prediction and attention networks can be used to study the relative importance \cite{Vaswani2017}. In this paper we use attention at each modality to understand the relative importance of the low-level or deep features within the modality. We also use attention layers while fusing the three modalities and learn the attention weights to find the ratios of importance of each modality. The paper by Querishi et.al. \cite{qureshi2019verbal} is the only paper closest to what we are doing in terms of using a subset of the dataset we are using and applying attention at one layer. By using multiple layers of attention at several levels we have been able to obtain much better results than them and the network is computationally less expensive due to attention operations, thus minimizing the test time of the framework.


\begin{figure*}
  \includegraphics[scale=0.56]{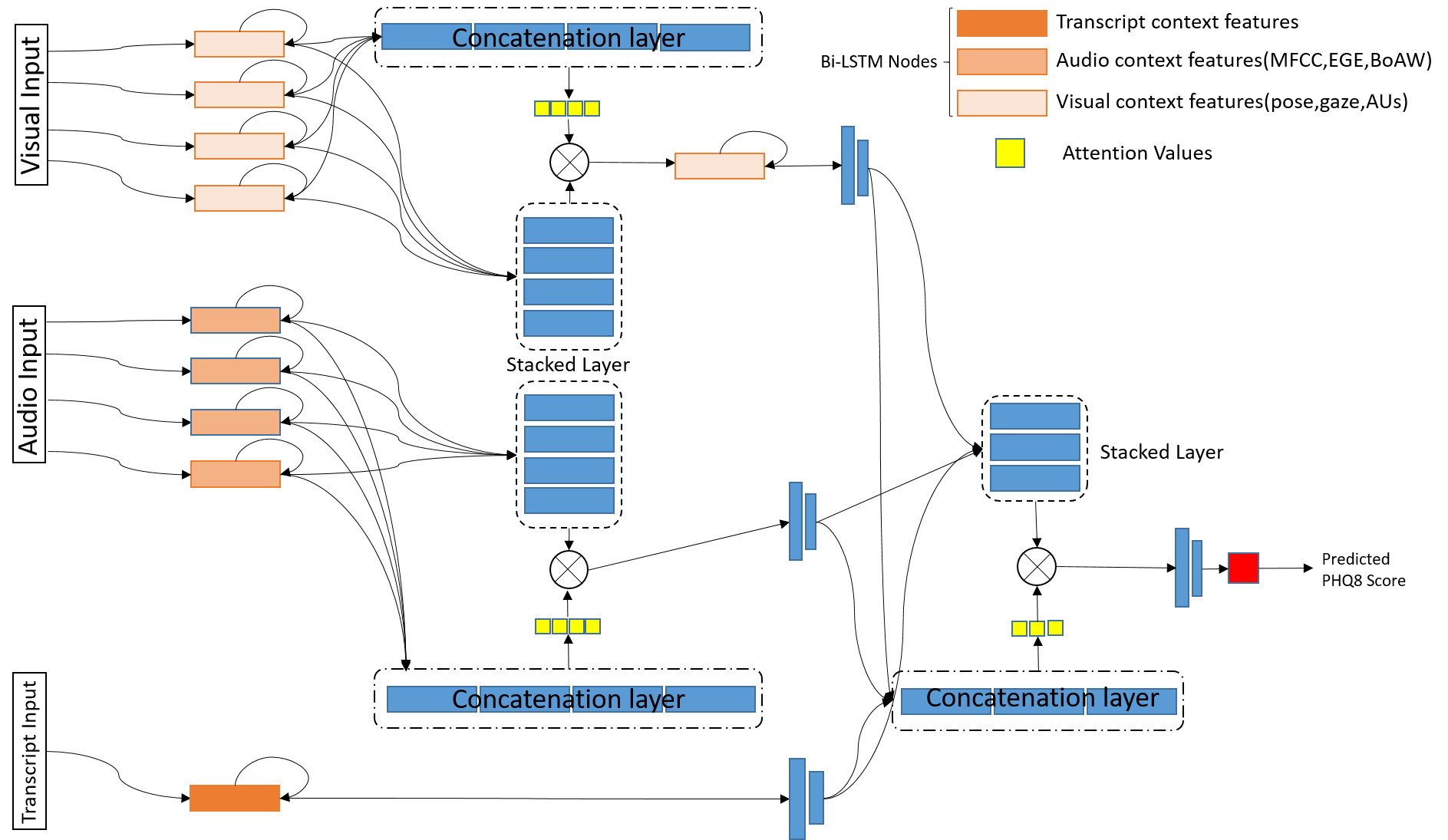}
  \caption{Block diagram of proposed multi-layer attention network on multi-modality input features}
  \Description{}
  \label{fig:Blockdiagram}
\end{figure*}

\section{Proposed Attention network for Hybrid fusion}
\label{sec:dataset}
A block diagram of the proposed multi-layer attention network is shown in Figure \ref{fig:Blockdiagram}. The attention layer over each modality teaches the network to attend to the most important features within the modality to create the context feature for that modality. The context features of each modality are passed through two layers of feedforward networks, and the outputs of these 3 feedforward networks are fused in another stacked BLSTM. The output of the 3 feedforward networks contain the most important features per modality and are fused to form another concatenation vector with an attention layer on top of it. The output of this attention layer is multiplied by the output of the stacked Bi-LSTM output and passed through the regressor. The loss of the regressor is back-propagated to train the weights learned at each level of the network ensuring end-to-end training.

\subsection{Dataset}
The Extended Distress Analysis Interview Corpus dataset (E-DAIC) \cite{DAIC} is an extension of DAIC-Wizard-of-Oz dataset (DAIC-WOZ) dataset. This dataset containing audio-video recordings of clinical interviews for psychological distress conditions such as anxiety, depression and post traumatic stress disorders. The DAIC-WOZ \cite{DAIC-Woz} was collected as part of a different effort which aimed to create a bot that interviews people and identifies verbal and nonverbal indicators of mental illnesses \cite{gratch2014distress} which has an animated virtual bot instead of a clinician and the bot is controlled by a human from another room. 
The Audio/Visual Emotion Challenge (AVEC 2019)  challenge \cite{AVEC2019} presents an E-DAIC where all interviews are conducted by the AI based bot rather than a human. This data has been carefully partitioned to train, development (dev) and test while preserving the overall gender diversity. There are total 275 subjects in the E-DAIC dataset out of which 163 subjects are used for train, 56 for both dev and test, out of which the test labels are not available as per challenge. Thus the results shown are mostly on the dev partition only.

\section{Methodology}
\label{sec:methodology}
In this section we describe the models created on each modality along with the various models created for fusion of different features from different modalities.

\subsection{Text modality}
We use the speech-to-text output for the participants in the data provided by \cite{AVEC2019}. Since several participants used colloquial English words, we modified the utterances by replacing such words with the original full word, otherwise they become all out of vocabulary words while training a neural network for language modeling or other predictions. We used pretrained Universal Sentence Encoder \cite{USE} to get sentence embeddings. To obtain constant size of the tensors, we zero pad shorter sentences and have a constant number of timesteps as 400. The length of each sentence embedding vector is 512 making the final array dimension as (400,512).
We used 2 layers of stacked Bidirectional Long short term memory network architecture with sentence embeddings as input and PHQ scores as output to train a regression model on the speech transcriptions. Each BLSTM layer has 200 hidden units, wherein the output of each hidden unit of the forward layer of first BLSTM layer is connected to the input of the forward hidden unit of the second layer. Same connections are built for each hidden unit in the backward layers as well to create the stacking. The two layers of BLSTM give an output of (batchsize,400) at each timestep and this is sent as an input to a feedforward layer for regression. We kept the number of nodes in the feedforward layers as (500,100,60,1) and used rectified linear units as the  activation function.

\subsection{Audio}
For the audio modality we created models using different audio features (low-level features as well as their functionals).
As functionals, the arithmetic mean and co-efficent of variations is applied on the low-level features and this is used as a knowledge abstraction on top of the low-level features \cite{Schuller2007TheRO}. The vocal  timbre is  encoded  by low-level descriptor features such as Mel-Frequency  Cepstral  Coefficients (MFCC)\cite{MFCCselection} and studies \cite{bone2014robust, MFCCselection} show that the  lower  order  MFCCs  are  more important for affect/emotion prediction and paralinguistic voice analysis tasks. 
The extended Geneva Minimalistic Acoustic Parameter Set (eGeMAPS) contains 88 features which include the GeMAPS as well as spectral features and their functionals. The GeMAPS feature consists of frequency related features (pitch, jitter, formant), energy related features (shimmer, loudness, harmonic-to-noise ratio), spectral parameters (alpha ratio, hammarberg ratio, Spectral slope -500 Hz and 500-1500 Hz, formant 1,2,3 relative energy), harmonic difference between H1-H2 and H1-A3, and their functionals, and six temporal features related to speech rate \cite{GeMAPS}. Apart from these low-level features mentioned above, a high dimensional deep representation of the audio sample is extracted by passing the audio through a Deep Spectrum and a VGG network. This feature is referred to as deep densenet feature in the rest of the paper.

For audio features, the span of vectors were the participant has spoken was only considered in our experimentation. 
Each of these features were available as a part of the challenge data and they have different sampling rates. The functional audio and deep densenet features are sampled at 1Hz, whereas the Bag-of-AudioWords (BoAW)  \cite{openXBOW} is sampled at 10Hz and the low-level audio descriptors are sampled at 100Hz. The length of the low-level MFCC features and low-level eGEMAPS is 39 and 23 respectively, and the total time steps is 140500 for both. For the functionals however, the lengths are 78 and 88 respectively with 1300 and 1410 timesteps. The BoAW-MFCC and the BoAW-eGEMAPS features are of length 100 and took 14050 timesteps for each. The deep densenet features are of 1920 dimension in length and took 1415 timesteps.

For the individual audio modality, we trained another stacked BLSTM network with two layers each having 200 hidden units. We take the last layer output and pass it to a multi-layer perceptron with each layer of (500,100,60,1) nodes in progression and Rectified Linear Units as activation function.

\subsection{Visual Modality} 
For the video features available in the challenge dataset, we experimented with both the low-level features as well as the functionals of the low-level video descriptors provided. We observed similar performance comparing the low-level descriptor with its functional. Since the deep LSTM networks could also learn similar properties from the data (like functionals and more abstract information), we chose to use the low-level descriptors as it has more information than its mean and standard deviation. Each low-level descriptor features for Pose, Gaze and Facial Action Units (FAU) are sampled at 10Hz. The length of these features were 6, 8, 35 respectively and all having 15000 timesteps. The Bag-of-VisualWords (BoVW) is also provided in the challenge data and has a length of 100 with 15000 timesteps. We use these features to train individual model per feature, all having a single layer of 200 BLSTM hidden units, followed by a maxpooling and then learn a regressor. We experimented with various combinations such as sum of all outputs, mean of outputs and also by maxpooling as three alternatives, but maxpooling worked best, so we have utilized maxpooling over the LSTM outputs.

\subsection{Fusion of Modalities}
Standard procedures of early fusion are computationally expensive and can lead to overfitting when trained using neural networks. Thus, late fusion and hybrid fusion models became more prevalent. We propose a multi-layer attention based network that learns the importance of each feature and weighs them accordingly leading to better early fusion and prediction. Such an attention network gives us insights of which features in a modality are more influential in learning. It also gives an understanding of the ratio of contribution of each modality towards the prediction.

Towards the fusion, we did several experiments within each modality and across. First, we fuse the low-level descriptors of the video modality. We take the gaze,pose, and facial action unit features and pass them through a single layer of 200 BLSTM cells and apply attention over them. The output of attention layer is passed through another BLSTM layer with 200 cells. We take global max pool of this LSTM output and pass through feed forward network with 128 hidden units. We call this fusion model as \textit{videoLLD-fused model} in table 2.
Second, we combine the low-level video features with the BoVWs and use a similar network of 200 hidden unit BLSTM layer followed by attention and another BLSTM, which is then passed through a feed-forward layer for regression. We call this fusion model as \textit{video-BoVW} fused in Table 2.

The third fusion model is created using the attention vector output from the video modality and the output of the text modality. These two outputs are combined and passed through a stacked BLSTM and an attention layer prior to the video regressor. This fused model is referred to as  \textit{Video-text fusion} in Table \ref{tab:perf2}.

The fourth fusion model uses audio and text modalities together. Again we took the output of the attention layer at each modality and build a hybrid fused network but passing them through two routes. In the first route the outputs of attention are concatenated and passed through attention followed by feed-forward layer and regression loss is propagated. Another route passes the output of both attention layers are passed through a stacked BLSTM of 2 layers each with 200 cells. An attention layer is applied on top of the stacked BLSTM layers and this output is fed to a feed forward network of 128 hidden units. This network is called \textit{Audio-Text fused} in Table \ref{tab:perf2} and is seen to have better performance than standalone audio model naturally due to use of text features which led to best results. 

Our fifth fusion model uses Video and text modalities together and here we again use attention layer at each sub-modality of the video inputs and then combine it with the text modality using another attention network over video and text. Quite surprisingly, the results of this fusion are very similar to audio and text modality fusion and the learning curve also ended up being quite similar. This network called the \textit{Video-Text fused} in Table \ref{tab:perf2} and, has one route that runs through a Bi-LSTM network of 200 units for each sub-modality and then for each time-step, they are fused together using an attention layer over all the incoming video features(Gaze, Pose, AUs and BOVW), which goes through another Bi-LSTM with 200 units to extract the contextual information from within the fused features.

Our sixth and final fused model uses all the modalities together. We use the attention based visual modalities to finally obtain a 128 unit vector, we use the attention based audio modalities to obtain a 128 unit vector and we extract the the information from the transcript modalities and derive a 128 bit vector from that. We again use another attention layer over these 3 modalities(Video, Audio and Text), to fuse them together and regress for the PHQ8 score. There were several challenges in integrating this fused model. We hypothesise that the error function consists of several local minima, which make it more difficult to reach the global minima. On testing the model with individual modalities, we observed that both Video feature model and audio feature model have a much steeper descent than the ASR model, on fusion, the model often got stuck on the minima of the video and audio features which are both quite close. To mitigate this and "nudge" the model towards a minima which takes the path of the minima reached by ASR transcripts, we multiply the final outputs of the attention layer element-wise with a variable vector initialised with values in the reciprocal ratios of the rmse loss for each individual modality in order to prioritize the the text modality initially. This led to a stable decline of train and validation loss, more stable than the individual modality loss also, and the final attention scores are indicative of contributions of each individual modality. Upon convergence, the attention ratios were [\textit{0.21262352, 0.21262285, 0.57475364} ] for video, audio and text respectively.



\begin{table*}
\centering
  \caption{Regression of PHQ score in terms of RMSE and MAE for each feature within each modality }
  \label{tab:perf1}
  \begin{tabular}{|c|ccccc|cccc|c|}
    \toprule
    \multicolumn{1}{c|}{\textbf{Partition}} &  \multicolumn{5}{c|}{\textbf{Audio}} &  \multicolumn{4}{c|}{\textbf{Video}} & \multicolumn{1}{c}{\textbf{Text}}\\
    \cline{2-11}
     & Funct MFCC&Funct eGeMAPS
&BoAW-M&BoAW-e &DS-DNet&Pose-LLD&Gaze-LLD&FAU-LLD&BoVW&Text\\
    \hline
    Dev-proposed & 5.11 & 5.52 & 5.66 & 5.50 & 5.65 &
5.85
& 6.13 & 5.96 & 5.70 & \textbf{4.37}\\
\hline
Dev-baseline \cite{AVEC2019} & 7.28 & 7.78 & 6.32 & 6.43 & 8.09 & - & - & 7.02 & 5.99 &- \\
\hline
Qureshi et.al. \cite{qureshi2019verbal} & - & - &-& -& -& 6.45 & 6.57&6.53&-& 4.80\\
\bottomrule
\end{tabular}
\end{table*}

\begin{table*}
\centering
  \caption{Different Fusion networks in terms of RMSE and MAE}
  \label{tab:perf2}
  \begin{tabular}{|c|ccccc|}
    \toprule
    \multicolumn{1}{c|}{\textbf{Partition}} &  \multicolumn{5}{c|}{\textbf{Attentive Fusion models}} \\
    \cline{2-6}
     & \textit{videoLLD-fused} & \textit{video-BoVW fused}
& \textit{Video-Text fused} & \textit{Audio-Text fused} & \textit{All-feature-fusion}\\
    \hline
    Dev-proposed & 5.55 & 5.38 &4.64 & 4.37 & 4.28\\
    \hline
Dev-baseline \cite{AVEC2019} & - & - & - & - & 5.03\\
\hline
    Qureshi et.al. \cite{qureshi2019verbal} & - &- & 5.11 & 4.64 & 4.14 \\
  \bottomrule
\end{tabular}
\end{table*}

\section{Results}
\label{sec:results}
This section presents the results of all regression models and their ablation studies in detail. The results of models trained on individual features from each modality as shown in Table \ref{tab:perf1}. We show results on 4 different types of fusion networks in Table \ref{tab:perf2}. Since the labels of the test data are not available as per the challenge, we show most results on the validation (dev) partition. The only results on test partition is from the text-based model with which we made a submission and got all scores from the challenge. The paper \cite{qureshi2019verbal} uses a subset of the E-DAIC data used by us which does not include the test partition and the dev partition could also be a little different, so we cannot directly compare the results but thats the closest comparison on similar dataset.

\subsection{Results from Text Modality}
The attention based BLSTM network trained on text transcriptions achieved best results in comparison to other modalities on the test set of both the E-DAIC (challenge data) and DAIC-WOZ dataset (current dev partition). This is in coherence with the observation of the clinicians that the verbal content is a significant marker and has explicit features which could influence the decision of depression stage classification. We achieved a root mean squared error (RMSE) of 4.37 on the development partition of the challenge dataset (as shown in Table \ref{tab:perf1}). 
we submitted the output of this model to the challenge, we have the detailed correlation co-efficients on the test partition only for the text modality and not for other modalities. On the test set, the text-based model is able to achieve a Mean absolute error (MAE) of 4.02, a RMSE of 4.73, which confirms to have a \textit{concordance correlation coefficient (CCC)} of \textbf{0.67} which is main metric in the challenge. The Pearson's Correlation coefficient  (PCC) of this model is 0.676, the coefficient of determination (r2) is 0.457, and the Spearman's Correlation coefficient (SCC) is 0.651 as per the results from the challenge on test set. Overall this network outperforms the state-of-art model \cite{qureshi2019verbal} by 8.95\%. The code converged at 15 epochs with a validation loss of 4.37 and batch size of 10 which was kept empirically. The average test time for a single text transcription to get predicted is 0.09 secs.

\subsection{Results from Audio Modality}
The loss in terms of RMSE on the audio features of the development split of the dataset given from the challenge is shown in Table \ref{tab:perf1} as \textit{Dev-proposed}. \textit{Dev-baseline} are the results provided in the baseline paper of the challenge \cite{AVEC2019}. In comparison with the baseline model, each of our individual networks outperformed in terms of RMSE. For the audio MFCC feature based model, we outperform baseline by 29.80\%, whereas for eGEMAPS we are better by 29.04\%. For the BoAW-MFCC we outperform by 10.44\% and in case of BoAW-eGE we achieve 14.46\% improvement over the baseline. Each individual audio feature code runs for 15 epochs with a batch size of 10. The average time required for one sample using functional MFCC is 0.23 secs, using Functional eGEMAPS is 0.14 secs, using BoAW-MFCC is 0.45 secs using BoAW-eGE is 0.45 secs and DS-DNet features is 0.13 seconds.
For the audio models, we tried a convolution neural network architecture for fusing the MFCC, eGEMAPS and DS-DNet features, but observed that the performance with Bi-LSTMs was slightly better than convolution networks due to the sequential learning capability which suits such features.

\subsection{Results from Visual Modality} 
The results on the video features are better than the baseline as well as the state-of-art, but is still worse than the results obtained on the text and speech modalities. Among the visual features, the Bag of Visual words performed best outperforming the baseline by 4.8\%. Comparing with \cite{qureshi2019verbal}, we outperformed them by 9.3\% using pose features, by 6.6\% using gaze and 8.7\% using facial action units.

\section{Results from Fused Modalities}
The details of the RMSE of each fusion model on the development partition is shown in Table \ref{tab:perf2} as Dev-proposed. The Dev-baseline are the results on the same split from the baseline paper. The third row which shows the results of the state-of-art paper is not on the same set, but on the test dataset of DAIC-WOZ. Assuming that the entire test partition of DAIC-WOZ is now a part of the validation/development partition of the challenge dataset, we present the comparison with them. The model using all features fused using multiple levels of attention led to the best results outperforming the baseline by 17.52\%. In comparison to \cite{qureshi2019verbal}, for the Audio-text and video text fusion networks, our networks are better by 5.8\% and 9.19\% respectively, but  our all-feature fusion network is slightly worse. This is not conclusive as the dataset being used in the paper is slightly different.
The attention mechanism automatically weighs each feature in each modality and allows the network to attend to the most important features for the regression decision. The network thus learns the relationship between the features with the PHQ-8 scores.


\section{Discussions and Future Work}
This paper proposes a multi-level attention based early fusion network which fuses audio, video and text modalities to predict severity of depression. For this task we observed that the attention network gave highest weights to the text modality and almost equal weightage to audio and video modalities. Giving higher weights to text modality is in coherence with clinicians as the content of speech is critical to diagnose depression levels. Audio and video are equally important sources of information and can be critical for the prediction of severity. Our intuition for lower importance to video data is the limited features that we could use from the video modality (eye-gaze, facial action-units and headpose). A clinician in a face-to-face interview can observe a person's body-posture (self-touches, trembling etc) or record electrophysiological signals, thus helping diagnose better. 

The use of multi-level attention led us to obtain significantly better results in all individual and fusion models compared to both the baseline and state-of-art. Using attention over each feature and each modality had a two fold advantage overall. Firstly, this gives us deep and better understanding of importance of each feature within a modality towards depression prediction. Secondly, attention simplified the network's overall computational  complexity and reduced the training and test time. Experimental results show that the model with all-feature fusion using multi-level attention outperformed the baseline by 17.52\%. The model built only on text modality was also significantly better in comparison to \cite{qureshi2019verbal} and also on the test set achieving a CCC score of 0.67 in comparison to 0.1 of the baseline on testset.

As future work, the authors want to be able to get rid of inductive bias from classes of data with more training samples and use few-shot learning techniques to be able to learn models with less data, as it is highly challenging to get more data or balanced data across classes in such domains. We also are trying to deep delve to understand what are the features that have a positive or negative influence in making these decisions between mild/severe depression. That would lead to more explainability in these models, which is of more importance to a clinician in understanding the output of these models.

\label{sec:conclusion}




\bibliographystyle{ACM-Reference-Format}
\bibliography{sample-base}



\end{document}